\documentclass[journal]{vgtc}                     % final (journal style)
\onlineid{0}

\vgtccategory{Research}

\vgtcpapertype{Applications}

\title{A Comparative Visual Analytics Framework for Evaluating Evolutionary Processes in Multi-objective Optimization}

\author{%
  Yansong Huang,
  Zherui Zhang,
  Ao Jiao,
  Yuxin Ma, \textit{Senior Member, IEEE},
  and Ran Cheng, \textit{Senior Member, IEEE}
}

\authorfooter{
    \item All authors are with the Department of Computer Science and Engineering, Southern University of Science and Technology, China. Y. Huang and Z. Zhang contributed equally to this work. Y. Ma is the corresponding author. Email: \{huangys2019,zhangzr32021,jiaoa2020\}@mail.sustech.edu.cn, \{mayx,chengr\}@sustech.edu.cn. 
}

\abstract{
  Evolutionary multi-objective optimization (EMO) algorithms have been demonstrated to be effective in solving multi-criteria decision-making problems. In real-world applications, analysts often employ several algorithms concurrently and compare their solution sets to gain insight into the characteristics of different algorithms and explore a broader range of feasible solutions. However, EMO algorithms are typically treated as black boxes, leading to difficulties in performing detailed analysis and comparisons between the internal evolutionary processes. Inspired by the successful application of visual analytics tools in explainable AI, we argue that interactive visualization can significantly enhance the comparative analysis between multiple EMO algorithms. In this paper, we present a visual analytics framework that enables the exploration and comparison of evolutionary processes in EMO algorithms. Guided by a literature review and expert interviews, the proposed framework addresses various analytical tasks and establishes a multi-faceted visualization design to support the comparative analysis of intermediate generations in the evolution as well as solution sets. We demonstrate the effectiveness of our framework through case studies on benchmarking and real-world multi-objective optimization problems to elucidate how analysts can leverage our framework to inspect and compare diverse algorithms.
}


\teaser{
  \centering
  \includegraphics[width=\linewidth]{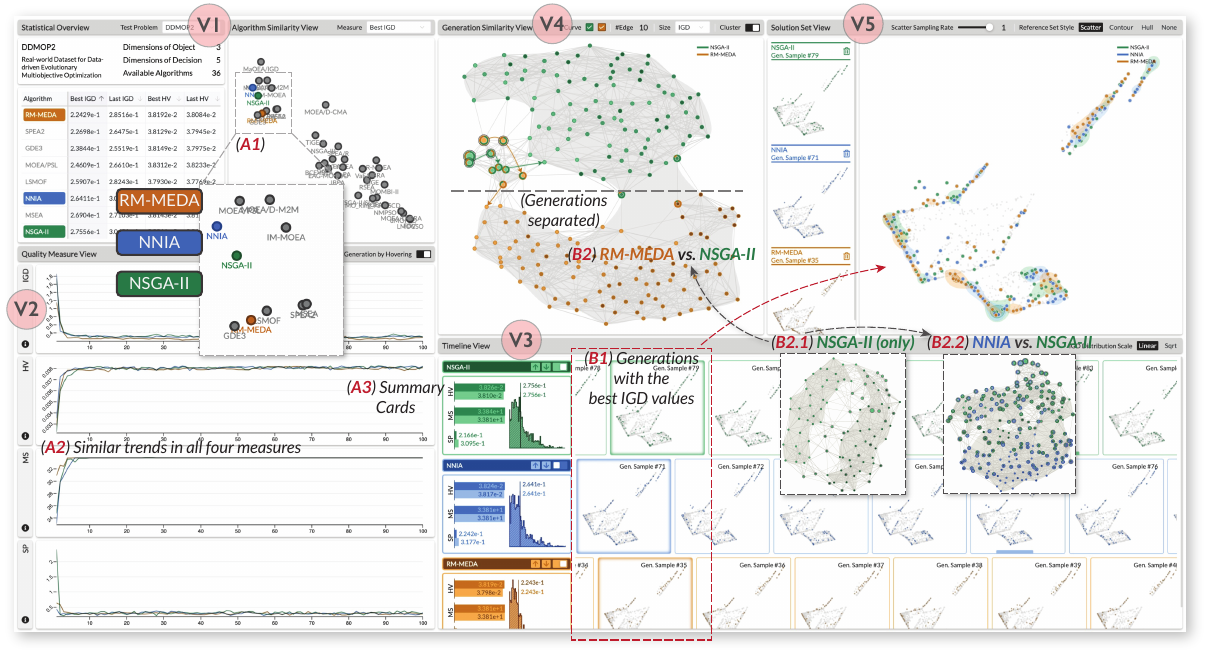}
  \caption{%
  Our visual analytics framework comprises three primary modules, namely the Algorithm-level Comparison module (V1), Evolution-level Exploration module (V2-4), and Solution-level Inspection module (V5). For the DDMOP2 test problem, the three algorithms show similar rankings and positions in the table and the projection result, respectively (A1), together with similar distributions on all four quality measures and the solution sets with the best IGD values in the respective algorithms (A2 and B1). The three $k$NN graphs display alternative relationships between generations from the same or different algorithms (B2, B2.1, B2.2).
  }
  \label{fig:teaser}
}

\graphicspath{{figs/}{figures/}{pictures/}{images/}{./}} % where to search for the images

\usepackage{inconsolata}

\usepackage{tabu}                      % only used for the table example
\usepackage{booktabs}                  % only used for the table example
\usepackage{lipsum}                    % used to generate placeholder text
\usepackage{mwe}                       % used to generate placeholder figures

\usepackage{mathptmx}                  % use matching math font
\usepackage{amsmath}

\usepackage{enumitem}

\usepackage[table,xcdraw,svgnames]{xcolor}
\definecolor{TODOcolor}{HTML}{FF9800}
\definecolor{AddContentcolor}{HTML}{0000CC}
\definecolor{REVISEcolor}{HTML}{FF0000}
\definecolor{classred}{HTML}{D73B36}
\definecolor{classorange}{HTML}{C16D28}
\definecolor{classblue}{HTML}{3768B1}
\definecolor{classgreen}{HTML}{167A48}
\definecolor{classpurple}{HTML}{8B79B8}

\newcommand{\REDCLASS}[1]{\textcolor{classred}{#1}}
\newcommand{\BLUECLASS}[1]{\textcolor{classblue}{#1}}
\newcommand{\ORANGECLASS}[1]{\textcolor{classorange}{#1}}
\newcommand{\GREENCLASS}[1]{\textcolor{classgreen}{#1}}
\newcommand{\PURPLECLASS}[1]{\textcolor{classpurple}{#1}}

\begin{document}

%%%%%%%%%%%%%%%%%%%%%%%%%%%%%%%%%%%%%%%%%%%%%%%%%%%%%%%%%%%%%%%%
%%%%%%%%%%%%%%%%%%%%%% START OF THE PAPER %%%%%%%%%%%%%%%%%%%%%%
%%%%%%%%%%%%%%%%%%%%%%%%%%%%%%%%%%%%%%%%%%%%%%%%%%%%%%%%%%%%%%%%


\firstsection{Introduction}

\maketitle
\vspace{-1.8mm}
Decision making under more than one criteria often appears in real-world optimization problems. Unlike single-aspect decision tasks where only one target will be satisfied, multiple aspects should be considered simultaneously to obtain optimal solutions with trade-offs between different yet often conflicting objectives. Considering water reservoir systems as an example~\cite{Reddy2006}, the operators are intended to design various operation policies that maximize power production. However, blindly optimizing the factors for power production may result in negative impact on irrigation or even increase the risk of flooding. To solve the multi-criteria decision making problems, various types of multi-objective optimization algorithms have been developed in the recent decades. The underlying mechanism of these algorithms is to search for a solution set where each solution cannot dominate the others, i.e., better than another solution on all objectives. Owing to the non-dominance nature in the solution set, decision makers are able to choose between a variety of feasible solutions to meet the incoming requirements.

Among the existing categories of methods, evolutionary multi-objective optimization (EMO) algorithms have been demonstrated as one of the most effective approaches to find an optimal solution set~\cite{Tian2021}. With a proper design of the evolution strategy, diversity and accuracy of solutions can be achieved simultaneously in a single run, providing wide-ranging decision choices for experts or stakeholders in real-world applications. Despite the fact that EMO algorithms are widely-used in many applications~\cite{sener2018multi,lu_neural_2022,abdolmaleki2020distributional}, a key challenge in evaluating EMO algorithms is to conduct comparative analysis between results from different algorithms~\cite{Zitzler2000,Li2018}. Conventional approaches utilize numerical quality indicators, such as Inverted Generational Distances (IGD) and Hypervolume (HV), to quantify the performance of the algorithms, and these metrics are naturally inherited in identifying the best or worst algorithms in comparison tasks. However, recent surveys~\cite{Li2019,Tian2021} emphasized that evaluating and comparing solution sets is a non-trivial task since aggregated measures, such as IGD and HV, are insufficient for characterizing EMO algorithms from multiple aspects including convergence, diversity, and uniformity. In addition, EMO algorithms usually work as a black-box, which could hinder the trustworthiness and in-depth evaluation of the behaviors in the evolutionary processes~\cite{Vincalek2021,YADAV2023}. Thus, there is an urgent need for incorporating comprehensive assessment in evaluating and comparing solution sets from different algorithms. 

Inspired by the success of visual analytics approaches in Explainable AI, we believe that the human-in-the-loop paradigm lends itself well in analyzing solution sets of evolutionary multi-objective optimization algorithms. In this work, we propose a visual analytics framework for comparative analysis of multiple EMO algorithms. The framework follows the algorithm-agnostic approach, allowing for the incorporation of various EMO algorithms as long as they meet the same evolutionary computing protocol. Supported by a multi-faceted visualization scheme, analysts are allowed to compare EMO algorithms at different levels of granularity, ranging from overall performance to individual generations. To gain insights into the evolutionary processes, a nearest-neighbor-based visual design is proposed to reveal the relationships between generations from multiple algorithms, which could benefit domain experts in understanding the underlying evolutionary behaviors of the algorithms. We demonstrate our framework through the widely-adopted DTLZ benchmarking suite and a real-world multi-objective optimization problem. Our contributions include:

\begin{itemize}[leftmargin=*]
    \item An interactive visual analytics framework for explaining and comparing the evolutionary multi-objective optimization algorithms;
    \item A suite of visualization and interaction designs that facilitate the exploration and comparison between measures, iterations, and individual solution sets;
    \item Case studies and interviews on analyzing evolutionary processes in two test problems to demonstrate the effectiveness of the framework.
\end{itemize}
\section{Related Work}

Our framework focuses on explaining and comparing evolutionary processes in EMO algorithms. In this section, we review the relevant works on visualization and visual analysis techniques in multi-objective optimization and algorithmic models.

\subsection{Visualization in Multi-objective Optimization}
\label{sec:background-mo}

Visualization has emerged as an effective means for analyzing solutions and attracted significant attention in the past decades~\cite{Pohlheim1999,Walker2013,Tusar2015}. Given the multi-dimensional nature of the decision and objective space, common techniques for visualizing high-dimensional data, including projections~\cite{Lotov2004,Talukder2020,Nagar2022} and parallel coordinates plots (PCP)~\cite{Li2017}, have been widely-adopted. Chen et al.~\cite{Chen2013a} propose a visualization design for characterizing Pareto fronts with self-organizing maps (SOM), while later work by Nagar et al.~\cite{Nagar2022} employs an enhanced, interpretable SOM that improves coverages and topographic correctness. Tusar and Filipic~\cite{Tusar2015} formulate the visualization problem of 4-D objective space as a multi-objective optimization problem, where the projected results should preserve shapes, ranges, and distributions of the objective vectors. Building on the 2-D RadViz plot, 3D-RadVis~\cite{Ibrahim2016-3dradvis} extends the ability to present data distributions of objective vectors by mapping the third dimension to the distances to a hyperplane. Meanwhile, PaletteViz~\cite{Talukder2020} proposes an alternative presentation by stacking multiple RadViz plots representing different layers based on the distances to a core location in the objective space. To address the readability issue of PCPs, Li et al.~\cite{Li2017} conducted a study on how PCPs reveal the distribution and quality of a solution set. 

Beyond static visualization techniques, interactive visual analysis methods have proven to be effective for exploring solution sets in an intuitive manner. Cibulski et al.~\cite{Cibulski2020} conduct a design study on how visualizing Pareto fronts can aid decision-making in the engineering field. They proposed an interactive system called PAVED that adopts a parallel coordinates plot to support the exploration of feasible solutions generated from optimization algorithms. More recent works in route planning~\cite{Weng2021} and interpretable machine learning~\cite{Chatzimparmpas2021} integrate the visualization of solution sets for a domain-specific problem into context-aware views, such as city maps with bus routes and multi-dimensional projections of trained models.

While various visualization approaches for solution set analysis have been proposed, there remains an urgent need for inspecting and exploring evolutionary processes, which may not be thoroughly facilitated by basic static visualizations of algorithms' final results. Our work supplements the visualization of solution sets and enables a more effective exploration of the dynamics and behaviors of evolutionary processes.

\subsection{Explainable AI}

Along with the recent advances in Explainable AI, there has been a considerable amount of research tackling the problem of explaining and diagnosing algorithmic models. Various surveys~\cite{Lu2017,Endert2017,Hohman2018,Guidotti2018blackbox,yuan2021survey,LaRosa2023,andrienko2021theoretical} have systematically summarized the research questions in this field. Here we mainly survey the literature related to visualizing execution processes of algorithms or models as well as model comparison.

\vspace{1.4mm}\noindent\textbf{Model Explainability.} Owing to the complexity of the underlying mechanisms, the black-box metaphor is widely-used in literature to describe the difficulties in understanding how a certain model works. Thus, some existing works tend to ``open the black box'' by exposing and interpreting the running processes of complex models. An early attempt proposed by Tzeng and Ma~\cite{Tzeng2005} depicts neurons and weights in an artificial neural network with a node-link-based design in order to show the dependencies between inputs and outputs. Muhlbacher et al.~\cite{Muhlbacher2014} provide a structured summarization of how visual exploration can be involved in an ongoing computational procedure. 

Regarding the execution stage involved in the explanation, Wang et al.~\cite{Wang2021} design CNN Explainer which supports education and inspection of the prediction processes. For analyzing training processes of convolutional neural network (CNN) models, DeepTracker~\cite{Liu2018a} addresses the challenge in visualizing large-scale training log data with a hierarchical down-sampling approach. Besides the CNN classification models, some works discuss the explainability issues in generative networks, such as Liu et al~\cite{Liu2018} and Kahng et al.~\cite{Kahng}.

\vspace{1.4mm}\noindent\textbf{Model Comparison.} Model comparison is a fundamental and critical task in assessing the performance of different models trained for a specific prediction problem~\cite{cawley2010over,raschka2018model,kotthoff2019auto}. In addition to selecting the best model or parameter setting, a thorough comparative analysis can also enhance the understanding of the learning and prediction behaviors. For typical classification tasks, comparisons between different classifiers can assist the understanding of the predicted class labels as well as the critical internal structures that affect the labeling process, such as Manifold~\cite{Zhang2018}, the ``learning-from-disagreement'' framework proposed by Wang et al.~\cite{Wang2022}, and the work by Gleicher et al.~\cite{Gleicher2020}. In graph learning, works by Pister et al.~\cite{Pister2021} and Xie et al.~\cite{Xie2022} focus on comparing important nodes or edges to disclose hidden patterns in network structures. With the outstanding success of deep learning methods in the past decade~\cite{Hohman2018,Choo2018,LaRosa2023,Wang2020a,Ye2022VISAtlas}, comparing neural networks becomes essential for disclosing learned knowledge in the complex models, including DeepCompare~\cite{Murugesan}, CNNComparator~\cite{Zeng2017}, VAC-CNN~\cite{Xuan2022}. Alongside the works mentioned above that explicitly compare neural networks, several advanced tasks also require model comparison as a critical component in their workflow~\cite{Wang2019atmseer,Ono2020,Ma2020,Wang2020,Yang2020}.

In summary, the visual analytics community has produced plenty of works that address the issue of explainability regarding computation processes in training or prediction stages as well as comparing model inputs and outputs. However, a research gap remains in the visual exploration of evolutionary processes which can help open the black box of EMO algorithms. To this end, PIVE~\cite{Kim2017} is by far the closest in spirit to our work, which illustrates a per-iteration visualization framework for iterative algorithms in machine learning and optimization. Nevertheless, adapting this framework to the analysis of EMO algorithms requires extra effort in addressing the domain-specific requirements in multi-objective optimization, including multi-aspect measures of the generations and support in understanding relationships between algorithms, evolutions, and solution sets.
\section{Design Overview}

In this section, we illustrate the fundamental components of evolutionary multi-objective optimization algorithms, which constitute the primary area of interest. Research challenges and analytical tasks are then outlined from a literature review and preliminary expert interviews.

\subsection{Background}
\label{sec:background-and-emo-intro}

\begin{figure}[t!]
	\centering	
	\includegraphics[width=0.99\columnwidth]{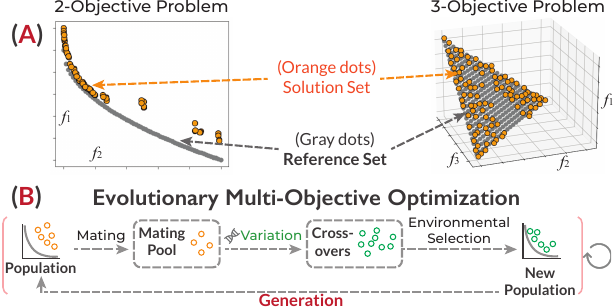}
	\caption{(A) Examples of the objective space for two and three-objective problems. (B) An illustration of typical evolutionary algorithm pipelines.}
    \label{fig:mop_example}
    \vspace{-4mm}
\end{figure}

\noindent\textbf{Definition and Terminology.} Given a decision space $X$, multi-objective optimization problems~\cite{Li2018,Tian2021} can be characterized as to find extremum of $m$ objective functions:
\vspace{-2mm}
\begin{equation}
\text{min}\; f(\mathbf{x}) = (f_1(\mathbf{x}), f_2(\mathbf{x}), ..., f_m(\mathbf{x}))
\vspace{-2mm}
\end{equation}

\noindent where $\mathbf{x} = (x_1, x_2, ..., x_d) \in X$ is a $d$-dimensional vector in the \textit{decision space}, and the space spanned by the objective functions forms an \textit{objective space}. Optimizing one objective function may often deteriorate the outputs of other objectives, making it almost impossible to find a single decision vector that minimizes all objectives. To formally represent relationships between solutions, we define \textit{Pareto dominance} as that for two solutions $\mathbf{x}_1$ and $\mathbf{x}_2$, $\mathbf{x}_1$ \textit{dominates} $\mathbf{x}_2$ if and only if $f_i(\mathbf{x}_1) \leq f_i(\mathbf{x}_2)$ and at least one $f_i(\mathbf{x}_1) < f_i(\mathbf{x}_2)$ (termed $\mathbf{x}_1 \prec \mathbf{x}_2$) on all $m$ objectives. Thus, we expect to find a set of trade-off solutions $\mathbf{x}'_1, \mathbf{x}'_2, ..., \mathbf{x}'_n, \in S$ where any two solutions in the set cannot dominate each other. All such $\mathbf{x}'_i$ in $S$ are denoted as \textit{Pareto optimal solutions}, and $S$ is thereafter called a \textit{Pareto set}. Accordingly, the corresponding objective vectors of $\mathbf{x}'_i$ in the objective space are called \textit{Pareto optimal (objective) vectors}, and the set of them forms a \textit{Pareto front}. \cref{fig:mop_example} (A) presents two examples of the solution distributions in two and three objective problems.

However, the majority of multi-objective optimization algorithms currently available can only offer an approximation to the ideal Pareto front. As a result, in benchmarking problems, a \textit{reference set} (gray dots in~\cref{fig:mop_example} (A)) is typically provided as a sampled representation of the continuous or discrete \textit{true Pareto front}~\cite{PlatEMO}. Various quality measures have been developed to quantify the proximity of the data distributions between a solution set and the reference set. These measures are then used to evaluate and compare the performance of different algorithms.

\vspace{1.4mm}\noindent{\textbf{Evolutionary Algorithms in Multi-objective Optimization.} The Evolutionary algorithm is a stochastic search strategy which has been shown to find optimal solutions that converge towards the ideal Pareto front while maintaining diversity in the solutions. \cref{fig:mop_example} (B) illustrates the typical pipeline of EMO algorithms, which involves the following main steps~\cite{Zitzler2004}. Initially, with a given population of decision vectors, a mating pool is created based on the individuals' fitness scores. Variations including recombination and mutation are then applied to the mating pool to create crossover individuals. In the final environmental selection step, the individuals from the original population and the modified mating pool are evaluated to build a new population via a survival testing strategy.
Such a complete loop of the steps is referred to as a \textit{generation}, and a new solution set is derived from the population in each generation. The entire process is executed iteratively until a termination criterion is met, such as thresholds for number of iterations or quality standards. As such, the solution set from the best generation can be treated as the final result of the algorithm.

It should be noted that strategies of mating, performing variations, and conducting environmental selection differ across EMO algorithms. For the purpose of supporting visual exploration and comparison of the evolutionary processes, we adopt an algorithm-agnostic paradigm based on generations. Under this approach, the evolutionary processes for different algorithms can be abstracted as a series of solution sets that correspond to the generations,~\cref{fig:overview}.

\subsection{Requirement Analysis}
Given the key features in evolutionary multi-objective optimization in~\cref{fig:mop_example} (B), we intended to design a visual analytics framework that incorporates comparisons between solution sets and generations in the evolutionary processes of multiple algorithms. To better illustrate the analytical tasks, we conducted a literature review on EMO algorithms~\cite{He2016,Li2019,Li2018,Tian2021} and compiled a list of major research gaps. In addition, we organized an open-ended pilot interview with two domain experts (E1 and E2) to validate and refine the research gaps. E1 (who serves as one of the co-authors) is a professor majoring in evolutionary computing and has 10+ years of research experience on EMO algorithms. E2 is a researcher in intelligent transportation systems who adopts multi-objective optimization algorithms in the workflow. 

During the interview, we discussed several questions with the experts, including their approaches to applying EMO algorithms in their research, whether they had used visualization methods in their workflow, and how interactive visualizations could facilitate their analysis. Both experts affirmed the value of visualization in inspecting and comparing solution sets when developing and benchmarking new algorithms. They also noted that many popular EMO tools~\cite{PlatEMO,Pymoo,geatpy} only provide basic scatterplots or PCPs of the solution sets with limited or even no interactions, highlighting the necessity for an interactive visual exploration framework. In addition, choosing an appropriate algorithm becomes a critical consideration when determining which algorithm is best suited for a given problem. During the development of new algorithms, comparative analysis plays an important role in uncovering whether these novel algorithms outperform existing counterparts. Specifically, we identified three key requirements for visual comparative analysis:

\begin{itemize}[leftmargin=*]\setlength\itemsep{0.1em}
    \item \textbf{R1: Level of details.} When it comes to identifying comparison targets~\cite{Gleicher2017}, it is important to consider three key components in EMO algorithms: quality indices, evolutionary processes, and individual generations. Quality indices are the most common way to compare the performance between algorithms and identify the best solutions~\cite{He2016,Li2019}. For an in-depth exploration of the algorithms, the evolutionary processes should be considered, which consists of the output of the generations in the iterative optimization process~\cite{Zitzler2000,Li2018}. E1 emphasized the value of investigating the behaviors of solution sets as they evolve iteratively, particularly when testing novel algorithms. In addition, comparing solution sets of individual generations facilitates a fine-grained quality analysis in the objective space.

    \item \textbf{R2: Patterns along the evolutionary processes or between individual generations.} In relation to the purpose (or ``actions'') of comparisons~\cite{Gleicher2017}, the experts have addressed two main categories of patterns in examining evolutionary processes: evolutionary patterns that entail comparing two (sub)sequences of generations, and patterns when comparing two individual generations. As surveyed in Section~\ref{sec:background-mo}, current research primarily targets on visualizing individual or the best generations, while exploring trends, progressions, and anomalies in the evolutionary process is an equally crucial aspect.
    
    \item \textbf{R3: Measures for assessing different quality aspects.} Regarding the comparison strategy~\cite{Gleicher2017}, the majority of research in EMO algorithms only reports limited types of quality measures for result comparisons, such as Inverted Generational Distance (IGD) and HyperVolume (HV). However, a more comprehensive analysis of solution sets in the evolutionary process necessitates a multi-aspect evaluation~\cite{Li2019,Tian2021}. In real-world multi-criteria decision-making scenarios, trade-offs between different quality aspects can enhance the diversity of the feasible solutions and offer a wider range of options to meet diverse requirements, as commented by E2.
\end{itemize}

\begin{figure}[!t]
	\centering	
	\includegraphics[width=1.00\columnwidth]{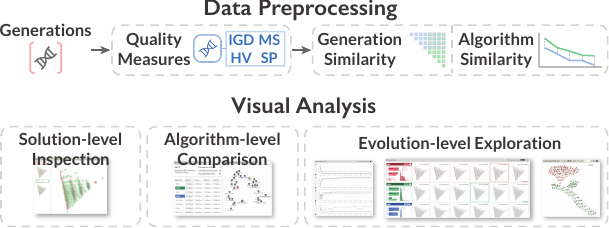}
	\caption{An overview of our visual analytics framework.}
    \label{fig:overview}
    \vspace{-4mm}
\end{figure}

\subsection{Analytical Tasks}
We further summarize the following tasks based on the aforementioned requirements. Notably, the first requirement, \textbf{R1}, is used as the primary axis to organize the tasks by aligning with the visual information-seeking mantra~\cite{Ben1996mantra} which provides the ``overview+detail'' scheme.

\vspace{1.4mm}\noindent\textbf{T1: Summarize the performance of algorithms.} Summarizing algorithm performance is the entry point of the comparative analysis. Analysts are interested in an overall comparison between algorithms:

\begin{itemize}[leftmargin=*]\setlength\itemsep{0.1em}
    \item \textit{How are the algorithms performed under different quality measures? What are the best solution sets in each algorithm?} (\textbf{R3})
    \item \textit{How similar are the algorithms with respect to best solution sets and quality measures?} (\textbf{R3})
\end{itemize}

\vspace{1.4mm}\noindent\textbf{T2: Reveal relationships among evolutionary processes.} At an intermediate level, the similarities between different evolutionary processes should be characterized to expose the behaviors of algorithms:

\begin{itemize}[leftmargin=*]\setlength\itemsep{0.1em}
    \item \textit{How do measures vary along the evolutionary processes?} (\textbf{R2}, \textbf{R3})
    \item \textit{How do solution sets in the generations change among different algorithms? How similar are such changes?} (\textbf{R2})
\end{itemize}

\vspace{1.4mm}\noindent\textbf{T3: Examine differences between solution sets.} At the generation level, a detailed comparison of solution sets from different generations should be considered as the fine-grained analysis:

\begin{itemize}[leftmargin=*]\setlength\itemsep{0.1em}
    \item \textit{How data in the solution sets distributes in the objective space?} (\textbf{R2})

    \item \textit{How similar are two solution sets in terms of different aspects of quality measures?} (\textbf{R3})
\end{itemize}
\section{Visual Analytics Framework}

Based on the identified considerations and analytical tasks, we have developed a visual analytics framework to support comparative analysis between EMO algorithms. Our framework consists of two main stages in the workflow:

\vspace{1.4mm}\noindent\textbf{Similarity Modeling.} As a data preprocessing stage, the logs recording the evolutionary processes of various algorithm candidates applied to a test problem are obtained and loaded into our framework. The similarities among the algorithms are subsequently computed, along with the similarities among solution sets of the corresponding generations.

\vspace{1.4mm}\noindent\textbf{Visual Exploration.} With the outputs from the preprocessing stage, we have designed three interactive visualization modules to inspect and compare the evolutionary processes from multiple granularities,~\cref{fig:overview}:

\begin{itemize}[leftmargin=*]
    \item \textit{Algorithm-level Comparison} (\textbf{T1}): The \textbf{statistical overview} shows the basic information of the test problem alongside quality measure statistics for all loaded algorithms. The overall similarity among all algorithms is visualized in the \textbf{algorithm similarity view}.
    
    \item \textit{Evolution-level Exploration} (\textbf{T2}): The \textbf{quality measure view} presents the trend of quality measures for all generations in selected algorithms, while the corresponding details are provided in the \textbf{timeline view}. The \textbf{generation similarity view} facilitates the exploration of inter-generational relationships between different algorithms.
    
    \item \textit{Solution-level Inspection} (\textbf{T3}): The \textbf{solution set view} enables direct comparison of data distributions in the objective space between the solution sets of selected generations.

\end{itemize}

\cref{fig:teaser} illustrates the interface of our framework. Analysts are allowed to switch between views to finish their tasks. Our modular design supports the loading of any EMO algorithm results on selected test problems once they follow the same output protocol.

\subsection{Data Preprocessing and Similarity Modeling}
\label{sec:preprocessing_stage}

The processing stage provides a formatted data abstraction of the evolutionary processes. Additional computations are performed including algorithm and generation similarities as well as quality measures on solution sets.

\vspace{1.4mm}\noindent\textbf{Data Abstraction and Sampling.} The evolutionary process for an algorithm can be regarded as a sequence of generations, each of which is associated with a solution set. As previously mentioned in Section~\ref{sec:background-and-emo-intro}, this data abstraction is commonly adopted by most EMO algorithms, allowing for algorithm-agnostic comparisons between evolutionary processes. However, the number of generations for an algorithm execution can be excessively large. As such, a uniform down-sampling strategy with a tunable sample rate can be applied in the preprocessing stage to reduce the size of the original generations before loading it, as illustrated in~\cref{fig:preprocessing} (A).

\vspace{1.4mm}\noindent\textbf{Quality Measures.} Our framework utilizes four distinct quality measures to evaluate solution sets with respect to key performance aspects including convergence, spread, and uniformity~\cite{Li2019} (\textbf{R3}),~\cref{fig:preprocessing} (B).

\vspace{1.1mm}\noindent\textit{Inverted Generational Distance (IGD)}~\cite{IGD2004}: IGD is one of the most widely-used measures in multi-objective optimization. Given a solution set $S = \{ \mathbf{x}_1, \mathbf{x}_2, ..., \mathbf{x}_n \}$ and a reference set $P^\ast = \{ \mathbf{r}_1, \mathbf{r}_2, ..., \mathbf{r}_m \}$, IGD is formulated as follows:

\begin{equation}
    \label{eq:IGD}
    \text{IGD}(P^\ast,\Omega)=\frac{\sum_{i=1}^m\min_{\mathbf{x} \in S}\text{dist}(\mathbf{r}_i,\mathbf{x})}{m}
\end{equation}

\noindent where $\text{dist}(\mathbf{r}_i,\mathbf{x})$ is the Euclidean distance from $\mathbf{r}_i$ to a solution $\mathbf{x}$. In other words, Equation~\ref{eq:IGD} calculates the average distance from each known ground-truth reference point $\mathbf{r}_i$ to its nearest solution in a given generation. A lower IGD value means a better performance of the solution set with respect to convergence towards the ground-truth as well as diversity on the true Pareto front.

\vspace{1.1mm}\noindent\textit{Hypervolume (HV)}~\cite{HV1998}: HV is another commonly-used quality measure for evaluating solution sets. By specifying an anchor point in the objective space, HV can be seen as the union volume of the hypercubes determined by the anchor point and each solution $\mathbf{x}$ in the solution set $S$. A higher HV value signifies a superior performance of the solution set.

\vspace{1.1mm}\noindent\textit{Spacing (SP)}~\cite{Spacing1995} and \textit{Maximum Spread (MS)}~\cite{Zitzler2000}: As per the taxonomy proposed in the study by Li et al.~\cite{Li2019}, IGD and HV are both classified as aggregated indicators that reflect multiple aspects through a single value. While IGD and HV can indicate the quality of convergence and diversity, the uniformity and spread of solutions is an equally significant aspect that needs consideration. To achieve this goal, we employ two additional quality measures, namely, SP and MS, to assess the uniformity and spread of the solutions and ensure a better representation of the entire Pareto front. SP aims to calculate the variation in distances between the solutions where a large deviation implies a non-uniform distribution, while MS gauges the covered area connected by the minimum and maximum values on each objective in the objective space.

\vspace{1.4mm}\noindent\textbf{Generation and Algorithm Similarity.} In EMO, the conventional approach for algorithm selection solely relies on quality measures as the basis for comparison. In this context, comparable measurement values are strong indicators to similar distributions of the solution sets and potentially similar evolutionary processes. For a more thorough comparison, we model the similarities in two detail levels for generations and evolutionary processes of algorithms, respectively (\textbf{R1}).

\vspace{1.1mm}\noindent\textit{Generation Similarity}: The purpose of assessing generation similarity is to quantify the degree of similarity between the distributions of solution sets in the objective space. To this end, we harness the Wasserstein distance, also referred to as Earth Mover's Distance (EMD), to reveal the similarity between the distributions,~\cref{fig:preprocessing} (C). The purpose of adopting this distance measure is that yielding varying solution sets is a common scenario in EMO algorithms, and the Wasserstein distance is effective in comparing distributions of varying shapes and sizes by measuring the ``work'' required to transform one distribution into another. Moreover, such interpretable and intuitive analogy of the distances, i.e., the amount of ``work'', facilitates the understanding for human users when conducting visual comparisons.

\vspace{1.1mm}\noindent\textit{Algorithm Similarity}: To analyze the similarities between algorithms, we consider the generation sequences in the algorithms as time series,~\cref{fig:preprocessing} (D), in order to utilize the well-established time series similarity measures. The attribute values for each time step can be assigned as the IGD or HV values, and we utilize the dynamic time warping (DTW) distance and the Euclidean distance to determine similarities between these time series. Additionally, we employ the generation similarity of the representative solution sets obtained from two distinct algorithms to represent their algorithm similarity, as evaluated by the generations possessing the best IGD or HV values.

\begin{figure}[!t]
	\centering	
	\includegraphics[width=1.00\columnwidth]{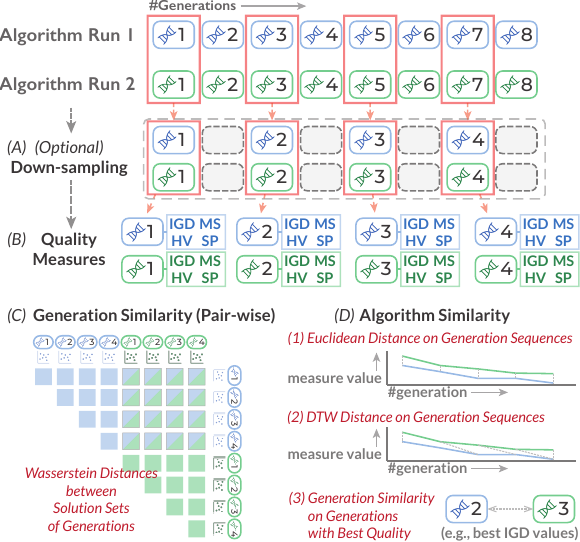}
	\caption{The data preprocessing stage. (A) An optional down-sampling stage to reduce the number of generations. (B) Four quality measures are calculated for each generation. (C) Generation similarity and (D) Algorithm similarity are measured between individual generations or the entire evolutionary processes.}
    \label{fig:preprocessing}
    \vspace{-4mm}
\end{figure}

\subsection{Comparing Algorithms}
The algorithm comparison module is designed to facilitate a coarse-grained analysis of the relationships between different algorithms by leveraging quality measures and similarities that were computed during the data preprocessing stage,(\textbf{T1}). Specifically, this module presents an overview of the quality measures and the interrelationships among the loaded algorithms on the test problem.

\vspace{1.4mm}\noindent\textbf{Visualization.} Shown in~\cref{fig:teaser} (V1), the region in the interface consists of three panels. In the statistical overview on the left side, the basic information regarding the workspace is presented, including the name of the test problem, the number of dimensions in both the decision space and the objective space, and the total number of algorithms loaded in the framework. A quality measure table is provided underneath the basic information, which itemizes the IGD and HV values for the best and last generations of each algorithm. 

To depict the coarse-grained similarities between the algorithms, the algorithm similarity view deploys a scatterplot to show the distances between algorithms with the t-SNE projection method~\cite{van2008visualizing} to determine the coordinates of the dots. Each dot on the scatterplot represents an algorithm, and the distance between algorithms is reflected in the relative positions of the dots. The input pair-wise similarity used for the t-SNE projection can be selected from the drop-down menu on the title bar of the view, where various algorithm similarity measures computed in the preprocessing stage are available for comparison.

\vspace{1.4mm}\noindent\textbf{Interaction.} Clicking on the algorithm names in the initial column of the quality measure table enables users to activate the visualization of the corresponding algorithm in the quality measure view and the timeline view. A categorical color scheme is applied to the activated rows with one color per each algorithm, whereby visual components sharing the same hue pertain to elements associated with the respective algorithm throughout the interface. In addition, analysts can click on the dots in the scatterplot to achieve the same effect as clicking on the algorithm names in the table.

\subsection{Analyzing Evolutionary Processes}
\label{sec:analyzing_evo_proc}

Since the behaviors of algorithms are manifested in the solution sets of the generations, analyzing the evolution of these generations can aid in comprehending the characteristics and underlying mechanisms of the algorithms. In this module, we present a set of visualizations that illuminate prominent evolutionary patterns (\textbf{T2}).

\vspace{1.4mm}\noindent\textbf{Quality Measure View.} Shown in~\cref{fig:teaser} (V2), this view is designed to illustrate overall temporal patterns of quality measures across evolutionary processes. It comprises four line charts, each corresponding to a specific quality measure, i.e., IGD, HV, MS, and SP. Upon activation of an algorithm in the quality measure table or the algorithm similarity view, the corresponding measurement values will be depicted as time series in the line charts. The horizontal axis of the line charts represents the order of generations, while the vertical ranges are scaled to fit the largest value in the visible area of the line charts. The colors of the lines are consistent with the colors assigned to the activated algorithms. The four line charts of the quality measures support zooming and panning, allowing for interactions while keeping the vertical value ranges filling the viewport of the charts. Analysts can select a specific generation in the timeline view by clicking on a data point in the corresponding time series.

\vspace{1.4mm}\noindent\textbf{Timeline View.} 
The timeline view,~\cref{fig:teaser} (V3), serves as a detailed explanation of the evolutionary processes for the activated algorithms. It shows the objective vectors of all generations in a juxtaposition manner, as illustrated in~\cref{fig:teaser} (V2). Each row represents an activated algorithm, consisting of two components.

\begin{itemize}[leftmargin=*]\setlength\itemsep{0.1em}
    \item \textit{Left Side}: A \textit{summary panel} at the beginning of the row displays the best measurement values of HV, MS, and SP in a bar chart. To provide a comprehensive view of the aggregated IGD value in Equation~\ref{eq:IGD}, we expand the representation measure with a histogram of all $\text{dist}(\mathbf{r}_i,\mathbf{x})$ in the definition. This allows analysts to observe the distributions of the nearest distances to the reference set and obtain a better understanding of the qualities of individual objective vectors in a solution set. The aggregated IGD value is marked with a vertical line. The value ranges in the bar charts and histograms share the same scale among all rows in the timeline view, enabling visual comparisons between multiple algorithms.
    
    \item \textit{Right Side}: A series of scatterplots are arranged as small multiples, with each scatterplot corresponding to a generation and presenting the distribution of the solution set in the objective space. If the number of objectives, i.e., the dimension of the objective space, exceeds two, PCA is adopted to project the objective vectors onto a 2-D plane. Note that we fit the PCA parameters based on the objective vectors in the reference set, and the derived projection matrix is shared among all the scatterplots in all algorithms. The dots in the scatterplots use the theme color of the corresponding algorithm, while the objective vectors in the reference set are plotted in gray color to depict a ground-truth.
\end{itemize}

The timeline view enables several interactions. By hovering over the bars in the bar chart, analysts can obtain the numerical values and the order of generations associated with each measure. Additionally, analysts may select a particular generation by clicking on the corresponding scatterplot, which triggers a mode change in the quality measures panel. In this comparison mode, the measures for the selected generation are displayed alongside those of the best-performing generations in a grouped bar chart style. Furthermore, to facilitate the comparison of the distributions of the best and selected generations, a histogram with stripe texture is drawn beneath the existing best-performing one, which represents the IGD distribution of the selected generation.

\vspace{1.4mm}\noindent\textbf{Generation Similarity View.} 
An issue that arises in the quality measure view and the timeline view is the lack of explicit depiction of relationships among individual generations. These relationships, whether they are related to the generations in the same or different algorithms, remain elusive in visual representations. To address this limitation, the generation similarity view, ~\cref{fig:teaser} (V4), has been introduced as a means of revealing similarities between generations, which aids analysts in identifying common or distinct evolutionary behaviors. \cref{fig:timecurve} (A) illustrates the design of this view. This approach allows for the inclusion of all generations of a given algorithm by toggling the switch adjacent to the algorithm name in the timeline view. Furthermore, it is possible to select multiple algorithms simultaneously.

\begin{figure}[!t]
	\centering	
	\includegraphics[width=1.00\columnwidth]{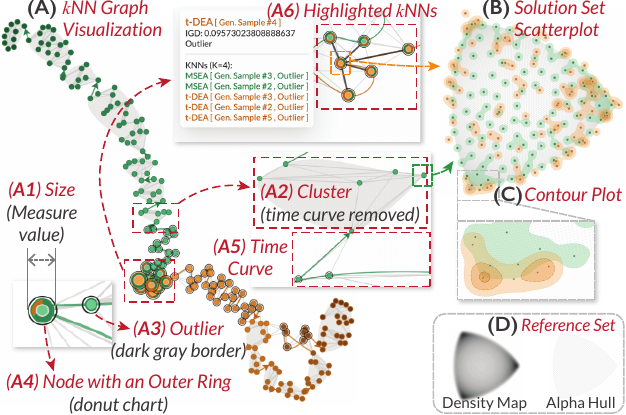}
	\caption{The visual design of (A) the generation similarity view and (B) the scatterplot in the solution set view.}
    \label{fig:timecurve}
    \vspace{-4mm}
\end{figure}

\vspace{1.1mm}\noindent\textit{KNN Graph Building:} To discern the underlying structure of the neighborhood in the chosen generations, we employ the $k$-nearest-neighbor graph ($k$NN graph) to delineate their proximity. The graph comprises nodes that correspond to generations from the selected algorithms, with nearest neighbors for each node detected based on the generation similarity, i.e., the Wasserstein distance between the corresponding solution sets in the objective space. Notably, the graph-building procedure includes generations from all selected algorithms, and this is critical for capturing the intra- and inter-relationships among the generations from different algorithms.

\vspace{1.1mm}\noindent\textit{Layout and Visual Encoding:} Once the $k$NN graph is prepared, we use the Kamada-Kawai layout algorithm~\cite{KAMADA19897} to compute the coordinates of the nodes, ~\cref{fig:timecurve} (A). The nodes are represented as colored points in the view, where the categorical color mapping for the activated algorithms is followed. In order to denote the sequential order of the generations, we assign varying degrees of lightness to the point colors with lighter colors indicating earlier positions in the evolutionary processes. Moreover, the size of the points can be mapped to one of the quality measures, which is controlled by a drop-down menu in the title bar of the view, ~\cref{fig:timecurve} (A1). The edges in the kNN graph are illustrated as light gray lines.

Beyond the backbone graph structure, we augment the visual presentation with several supplementary design elements to facilitate the identification of significant structures or features hidden in the graph.

\begin{itemize}[leftmargin=*]\setlength\itemsep{0.1em}
    \item While the $k$NN graph is useful for showing the neighborhood structure of generations, it may not accurately convey the intrinsic clustering patterns in the measurement space. To address this, we use the HDBSCAN~\cite{mcinnes2017accelerated} clustering algorithm with the pair-wise generation similarities as input to detect groupings and outliers among the generations. The nodes belonging to the same cluster are enclosed by a gray bubble underneath, ~\cref{fig:timecurve} (A2). Furthermore, the outlying nodes identified by HDBSCAN are decorated with a dark gray border, indicating that they do not belong to any clusters,~\cref{fig:timecurve} (A3).

    \item Nodes with nearest neighbors from different algorithms may suggest that the corresponding evolutionary processes share similar behaviors and produce comparable solution sets. We signify such nodes through a nested design where an additional outer ring is included,~\cref{fig:timecurve} (A4). This outer ring is a colored donut chart and represents the proportions of neighboring nodes that originate from the same or different algorithms.

    \item It may be sometimes challenging to trace the sequential ordering of generations based solely on the lightness of the node colors. Inspired by the Time Curve technique from Bach et al.~\cite{Bach2016}, we use a B\'ezier curve to emphasize the temporal sequence of nodes from the same algorithm, ~\cref{fig:timecurve} (A5). However, the curves may result in severe visual clutter with a large number of nodes. To address this, the curves are partitioned into segments by removing curves that lie within a cluster (A2 in~\cref{fig:timecurve}). In this manner, only the connections between clusters are displayed, which significantly reduces the visible length of the curves while preserving chronological information.
\end{itemize}

\vspace{1.1mm}\noindent\textit{Interactions:} A rich set of interactions are supported in the generation similarity view. Semantic zooming is supported in exploring the details of the $k$NN graph. The gray bubbles, outer rings, and time curves can be disabled to provide a clear interface. To avoid visual clutter, the number of visible $k$NN graph edges is controlled in a slider. Hovering the mouse pointer on the nodes and time curves opens an information prompt that contains details such as node neighborhoods or covered nodes of the curve, ~\cref{fig:timecurve} (A6). Clicking on a gray bubble can highlight the nodes in the cluster. 

\vspace{1.1mm}\noindent\textit{Alternative Design}: We have also considered utilizing chronological order as the primary horizontal axis to layout the nodes and representing similarities with edges between generations, which is similar to the CareerLine design~\cite{Wang2022seekfor}. Nonetheless, when the number of generations is large, such edges may result in severe visual clutter. Moreover, we prioritize similarity information between generations in the layout, which can be naturally encoded as 2-D distances.

\subsection{Inspecting Solution Sets}
Upon examining the evolutionary processes based on quality measures and generation similarities, a nuanced comparative analysis of particular generations may be required. Thus, we design a solution set view that enables a direct comparison of the solution sets in the objective space.

\vspace{1.4mm}\noindent\textbf{Visualizing Solution Sets.} Illustrated in~\cref{fig:teaser} (V5), when analysts select a particular scatterplot patch in the timeline view, the corresponding scatterplot is magnified and displayed on the right side of the solution set view (~\cref{fig:timecurve} (B)), with the relevant patch snapshot listed on the left side. The coordinates of the solutions in the generations follow the same approach described in Section~\ref{sec:analyzing_evo_proc}, wherein the projection matrix of PCA is fit on the objective vectors in the reference set and then applied to the selected generations. The colors of the dots are determined based on their originating algorithms. To mitigate potential visual clutter, we employ two additional methods:

\begin{itemize}[leftmargin=*]
    \item The first additional design is to draw a contour map for each selected generation using the kernel density estimation method, ~\cref{fig:timecurve} (C). This is intended to reveal the density of the dots, and the density levels are represented by the lightness of the filled contour colors and the stroke width of the contour lines.

    \item The second design is the application of the outlier-biased random sampling method~\cite{Liu2017b,Yuan2021Evaluation} to reduce the number of visible dots while maintaining the coverage of the overall data distributions. However, this sampling method is applied to the projected 2-D plane, which may lose outliers in the original objective space. To address this issue, the Local Outlier Factor method~\cite{breunig2000lof} is employed in the objective space to preserve all outlying extremum solutions. The detected outliers and extremum solutions are marked as crosses instead of dots, thereby maintaining the spread information of the solution sets.
\end{itemize}

\vspace{1.4mm}\noindent\textbf{Visualizing the Reference Set.} To establish a ground-truth for comparison, the reference solutions in the reference set are also visualized in the scatterplot. Due to the excessively large quantity of reference solutions, we design three different display modes that highlight various aspects of the true Pareto front. In~\cref{fig:timecurve}, the reference solutions can be presented as scatters of their objective vectors (B), a grayscale density map, or a textured gray alpha hull that depicts the boundaries of the reference set (D). Analysts can switch between the three modes flexibly, depending on the desired aspect of information they wish to examine, such as density for the density map and coverage for the alpha hull.

\vspace{1.4mm}\noindent\textbf{Interactions.} The scatterplot supports semantic zooming to facilitate detailed examination on demand. An exclamation mark displayed in a generation snapshot indicates that certain solutions lie outside the current viewport, prompting analysts to zoom out to locate them. Hovering over the dots opens a tooltip of the objectives' actual values. A slider is used to control the sample rate of the sampling method.
\section{Case Study and Expert Interview}

In this section, we describe how our framework can facilitate the understanding and investigation of the interrelations among EMO algorithms on a test problem. Furthermore, we outline the feedback from domain experts in an interview. Our framework follows a browser-server architecture and relies on the PlatEMO~\cite{PlatEMO} platform for running EMO algorithms and generating logs of the generations, including the solution sets for each generation. Python Flask library is employed for the backend, while Vue3 and d3.js are used for the frontend.

\subsection{DTLZ3 Benchmark Problem}
\label{sec:case1}
The DTLZ (Deb-Thiele-Laumanns-Zitzler) suite is a family of multi-objective optimization test problems initially proposed by Deb et al.~\cite{Deb2002DTLZ}, which is widely-used in evaluating the performance of EMO algorithms. In each problem, a known Pareto front is provided in the form of a reference set, enabling researchers to assess the accuracy and diversity of the approximations obtained from EMO algorithms. The first case study aims to investigate the ability of selected EMO algorithms to tackle the DTLZ3 problem, which involves three objective functions and a 12-dimensional decision space. A total of 36 EMO algorithms\footnote{Please refer to the supplemental material for a full list of the algorithms.} were utilized to test the DTLZ3 problem with the same initial population size of 100. We run each algorithm once with 500 generations and down-sampled them to 100 uniformly. The parameter $k$ for the $k$NN graphs in the generation similarity view was set to 10.

\vspace{1.4mm}\noindent\textbf{Algorithm-level Comparison (T1).} After loading data into the framework, the quality measure table and algorithm similarity view are displayed. To filter the algorithms with optimal performance, the table is ranked based on the generations with the best IGD values in each algorithm, i.e., the ``Best IGD'' column, in ascending order. It can be observed that \GREENCLASS{t-DEA}, \BLUECLASS{MSEA}, \ORANGECLASS{FDV}, \REDCLASS{NSGA-III}, and \PURPLECLASS{VaEA} are the most effective algorithms in the table. After activating these five algorithms, we find that their corresponding projected points were clustered together when switching to the ``Best IGD'' measure in the algorithm similarity view, consistent with the rankings in the table.

When the algorithm similarity view was switched to ``Euclidean (IGD)'' which measures the IGD value series across generations, FDV and MSEA were found to be significantly distant from the other three. This implies that they present different behaviors related to the changes in the IGD values during evolutionary processes. This observation is also confirmed by examining the IGD line chart in the quality measure view, which shows significantly lower values in the first ten generations for the two corresponding time series of FDV and MSEA. The HV values indicate similar trends, with FDV and NSGA-III showing outlying behaviors compared to t-DEA, MSEA, and VaEA.

\begin{figure*}[h!]
	\centering	
	\includegraphics[width=1.99\columnwidth]{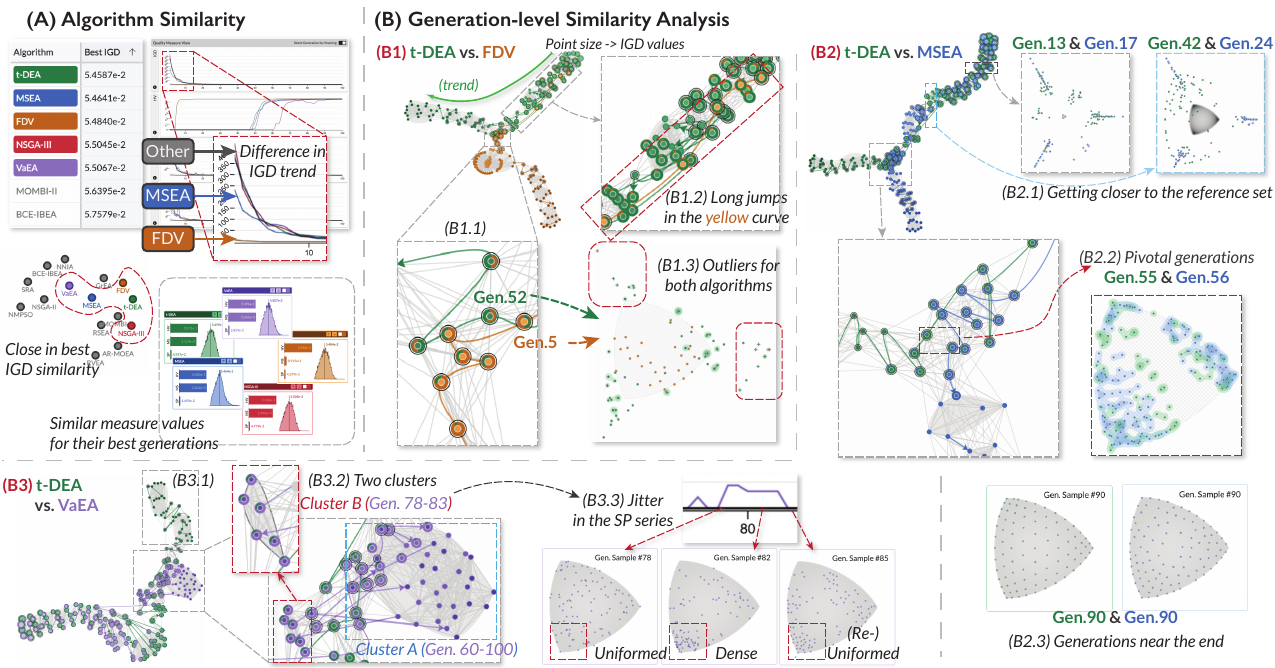}
	\caption{The DTLZ3 problem. (A) The algorithm-level similarity analysis. (B) Comparative analysis between t-DEA and other three algorithms.}
    \label{fig:case1}
    \vspace{-5mm}
\end{figure*}

\vspace{1.4mm}\noindent\textbf{Generation-level Similarity Analysis (T2, T3).} To comprehensively investigate the five activated algorithms, we utilize the interplay between various views to analyze the evolutionary processes. We observe from the summary panels in the timeline view that the activated algorithms show similar best measure values and IGD value distributions. However, due to the differences in the HV and IGD series mentioned above, we need further investigation (\textbf{T2}). To this aim, we select t-DEA, which has the best IGD value in the quality measure table, as the standard and perform paired comparisons with the other activated algorithms in the generation similarity view.

\vspace{1.1mm}\noindent\textit{\textbf{\GREENCLASS{t-DEA}} vs. \textbf{\ORANGECLASS{FDV}} (\cref{fig:case1} (B1))}: In the $k$NN graph constructed using the generations obtained from t-DEA (red) and FDV (blue), we make the following observations (\textbf{T2}). First, both algorithms depict clear evolutionary trends from the top right corner to the bottom left. When setting the visual mapping of point sizes to IGD values, it can be observed that the sizes reduce gradually along the evolution paths. Furthermore, there are only a few intersections between them, ~\cref{fig:case1} (B1.1). This implies that the two algorithms are designed to have distinct evolutionary strategies with little to no overlap. Second, in the intersection area, FDV's generations are in the early stages of the entire process (from generation \#5 to \#7), whereas t-DEA's process is in the intermediate stage (from \#51 to \#54). This suggests that FDV converges rapidly to the Pareto front and achieves a quality that t-DEA requires over 50 generations to reach. This pattern is more evident when enabling the time curves, where long jumps in the yellow curves are observed from the top-right corner,~\cref{fig:case1} (B1.2). We further examine the solution sets of generation \#5 in FDV and \#52 in t-DEA (\textbf{T3}), ~\cref{fig:case1} (B1.2). The scatterplot of the solution sets indicates that both generations have outlying solutions beyond the reference set area, and their similarities are relatively larger than those of the best generations, which is the main cause of having such intersections.

\vspace{1.1mm}\noindent\textit{\textbf{\GREENCLASS{t-DEA}} vs. \textbf{\BLUECLASS{MSEA}} (\cref{fig:case1} (B2))}: The $k$NN graph for t-DEA and MSEA exhibits a heavy overlap until almost half of the evolutionary process (over 50 generations) (\textbf{T2}). In the overlapping early stages, the solution sets of both algorithms are far from the reference sets but have the momentum to converge towards the central area, as depicted in the solution set scatterplots in~\cref{fig:case1} (B2.1). The evolutionary processes begin to diverge after approximately 50 generations, with the later generations forming two distinct branches. When comparing the solution sets of the pivotal generations (\#55 in t-DEA and \#56 in MSEA) and their subsequent generations, ~\cref{fig:case1} (B2.2), we observed that t-DEA achieves a more uniform coverage to the reference set, whereas MSEA presents poorer uniformity near the central area (\textbf{T3}), ~\cref{fig:case1} (B2.3),.

\vspace{1.1mm}\noindent\textit{\textbf{\GREENCLASS{t-DEA}} vs. \textbf{\PURPLECLASS{VaEA}} (\cref{fig:case1} (B3))}: During the early stages of the evolutionary process, there is an interweaving pattern between the two algorithms (\textbf{T2}). Similar to the comparison between t-DEA and MSEA, two diverging branches begin to emerge after 50-60 generations. However, the subsequent development behaviors differ between the two algorithms. As depicted in~\cref{fig:case1} (B3.1), the generations in t-DEA continue to evolve till the end of the process. Nevertheless, in~\cref{fig:case1} (B3.2), an intriguing cluster A is observed where the points with varying color lightness are mixed up. By observing the points through hovering, it can be determined that they correspond to generations ranging from \#60 to \#100. After enabling the time curve for VaEA, multiple connections among cluster A, cluster B, and some outlying generations in between are observed. Further examination reveals that the points in cluster B correspond to generations from \#78 to \#83. By analyzing the line charts in the quality measure view, jitters are observed between the 77th and 83rd generations in the time series of VaEA,~\cref{fig:case1} (B3.3). We further scrutinize the small multiples of scatterplots of VaEA in the timeline view from the 70th to 90th generations (\textbf{T2, T3}). In the ranges where jitters occur, some solutions start to move towards the bottom corner and then spread away after the 86th generation, indicating a sudden exploitation behavior in the evolutionary process.

\vspace{1.4mm}\noindent\textbf{Experts' Comments.}
First, FDV displays significantly faster convergence compared to other algorithms. With such observation, the experts mentioned that further investigations could be conducted to determine whether such a phenomenon is due to the inherent advantages of FDV or merely a coincidental matching between FDV's customized reproduction operators and the objective functions of the test problem. This is crucial in directing the development of more generalized algorithm designs rather than those tailored to specific test problems. Second, VaEA's behavior aligns with their knowledge that ``novelty search''~\cite{mouret2011novelty} can be initiated in evolutionary processes when optimization encounters barriers. The experts commented that despite acknowledging the presence of such behavior before using our system, it remains essential to employ intuitive visualizations to observe when and how novel solutions can be generated. This may also assist in the future optimization of variation strategies to produce superior novel solutions.

\subsection{DDMOP2 Problem}
\label{sec:case2}
DDMOP~\cite{He2020} is a test suite that were developed from real-world multi-objective optimization problems. In the second case study, we have selected the DDMOP2 test problem, which involves generating optimal frontal structures for vehicles to handle crashes. The problem consists of five decision variables for reinforced members around the frontal structure. Three factors representing the severity of a crash are assigned to the three objectives, which need to be minimized. We have employed the same algorithm and parameter settings as the first case study.

\vspace{1.4mm}\noindent\textbf{Algorithm-level Comparison (T1).}
Our exploration commences with the \GREENCLASS{\textbf{NSGA-II}} algorithm, which is the baseline employed in the reference~\cite{He2020}. We have employed \BLUECLASS{\textbf{NNIA}} and \ORANGECLASS{\textbf{RM-MEDA}}, which are the most similar and the best algorithms, respectively, based on the algorithm similarity of the generations with the best IGDs,~\cref{fig:teaser} (A1). After examining the line charts in the quality measure view (\cref{fig:teaser} (A2)) as well as the summary panel (\cref{fig:teaser} (A3)) in the timeline view, we have observed that the three algorithms exhibit a similar trend in the evolutionary processes and achieve comparable measurement values. However, it is still necessary to determine whether the evolving behaviors concealed in the generation series are identical.

\vspace{1.4mm}\noindent\textbf{Generation-level Similarity Analysis (T2, T3).}
Initially, we enabled the best generations of the three algorithms on IGD values in the solution set view,~\cref{fig:teaser} (B1). In the scatterplot, the three solution sets appear to display a similar distribution, which further supports the similarity observed in the quality measures (\textbf{T3}). However, evolutionary processes in the generation similarity view indicate differences in comparisons. As depicted in~\cref{fig:teaser} (B2), when only the $k$NN graph for NSGA-II is enabled, two major clusters appear, with intra-connections established by several outlying generations between the two clusters. After adding the generations from NNIA, a considerable number of nodes are displayed with outer rings, indicating that the generations from different algorithms are interconnected in the kNN graph significantly (\textbf{T2}). However, when comparing generations from NSGA-II (\cref{fig:teaser} (B2.1)) and RM-MEDA (\cref{fig:teaser} (B2.2)), the nodes are clearly separated based on the originating algorithms, with only a few inter-algorithm neighboring relationships preserved.

\vspace{1.4mm}\noindent\textbf{Experts' Comments.} For relationships between the three algorithms, domain experts observed that this case exemplifies a situation where algorithms that belong to the same category may not display substantial differences in evolutionary behaviors. NSGA-II and NNIA share the concept of non-dominated neighbor-based selection in the population selection stage, as described in previous research~\cite{NNIA2007}. Although updated crowding-distance measurements in NNIA have been reported to provide advantages in certain scenarios, additional inspection should be conducted to make sure whether the pronounced similarity between NSGA-II and NNIA results from the DDMOP2 test problem settings, which may not effectively showcase such advantages. This also highlights the issue that in real-world test problems like DDMOP, characteristics that maximize the advantages of a particular algorithm may not be as apparent as in artificial benchmarking problems, which presents new challenges in real-world decision-making scenarios.

\subsection{Expert Interview}

We conducted an expert interview involving three domain experts to further evaluate our framework. E1 and E2 were the experts who have participated in a preliminary interview for compiling analytical tasks, while E3 is a research scientist with a Ph.D. in evolutionary optimization and machine learning. During the interview, we presented the background, analytical tasks, and visualization design of the framework as well as the two test problems used in the case study. We then facilitated an open discussion, allowing the experts to freely explore and provide feedback on the implemented system. Comments on the framework design and functionality were collected during the interviews, which lasted between 1 and 1.5 hours each. Some of the comments provided by the experts on the case study results have already been reported in ``Experts' Comments'' in Section~\ref{sec:case1} and~\ref{sec:case2}.

All three experts agreed on the effectiveness of the proposed visual analytics framework for comparative analysis of EMO algorithms. E3 noted that such interactive comparison had not been extensively explored in the field of evolutionary computing and multi-objective optimization, thus highlighting the usefulness of the framework in terms of revealing the relationships between different algorithms. E1, who has experience in designing EMO algorithms, remarked that the framework could prove to be helpful in inspecting the effectiveness of the newly proposed variation and environmental selection strategies based on existing baselines. ``Sometimes, when designing new algorithms for specific types of problems, it is uncertain to what extent the strategy will work, and existing quality measures may not reveal the fundamental differences between two algorithms. An intuitive comparison could facilitate the assessment of whether the algorithm is performing in accordance with the researcher's expectations.''

With respect to the visualization design in the framework, positive feedback was received from the domain experts. E2 noted that although some existing EMO platforms and libraries have incorporated visualization features, an integrated visualization framework with comprehensive support for visual exploration could alleviate the burden of programming. In addition, the interplay between multiple views facilitates the understanding of evolutionary processes from diverse perspectives. E1 and E3 also highlighted the generation similarity view as an interesting approach to understanding the similarities between generations. ``It allows us to easily perceive the characteristics of various evolutionary processes, such as how fast an algorithm can converge and whether unexpected exploratory evolutions, such as Cluster B in the result of VaEA on DTLZ3, occurred,'' commented E1.

We also received constructive suggestions to improve the current framework design. E2 discussed the feasibility of integrating application-specific visualizations into the framework to provide better support for real-world decision-making scenarios. E3 suggested exploring the evolutionary processes in single-objective optimization problems, which would be an interesting topic for further research.
\section{Discussion and Conclusion}

In this paper, we present a visual analytics framework for comparative analysis of multiple evolutionary multi-objective optimization (EMO) algorithms. Based on the analytical tasks, the algorithm-agnostic framework allows for conducting comprehensive comparison from multiple levels of details including the algorithm-level assessment, evolution-level comparison, and solution-set-level inspection. The effectiveness of the proposed framework is demonstrated through two test problems.

\vspace{1.4mm}\noindent\textbf{Comparison with Existing Work.} To better contextualize our framework within the current literature on visualizing multi-objective optimizations, we have chosen three representative works and compared them across various dimensions in a qualitative manner,~\cref{fig:compare}. Our framework offers an interactive solution primarily concentrating on the analysis of evolutionary processes, whereas the existing works mainly tackle the task of solution set analysis for individual generations.

\begin{figure}[!t]
	\centering	
	\includegraphics[width=1.00\columnwidth]{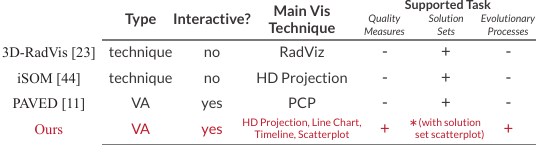}
	\caption{A comparative analysis involving 3D-RadViz~\cite{Ibrahim2016-3dradvis}, iSOM~\cite{Nagar2022}, PAVED~\cite{Cibulski2020}, and our framework is presented. In the columns denoting supported tasks, + symbolizes the primary task supported by the work, * designates partially or indirectly supported tasks, and - means tasks that are not supported.}
    \label{fig:compare}
    \vspace{-4mm}
\end{figure}

\vspace{1.4mm}\noindent\textbf{Scalability.} We discuss the scalability of our framework from two aspects: data preprocessing, and visual design.

\vspace{1.1mm}\noindent\textit{Data Preprocessing}: A bottleneck that impedes the scaling up of the number of algorithms is attributed to the similarity computation process. The computation of the Wasserstein distance incurs a high computational cost due to the approximation required in its computation~\cite{feydy2019interpolating}. Furthermore, the time required for pairwise similarity computation increases substantially with the number of generations in each algorithm. In the two case studies presented in this paper, the data preprocessing stage for each problem with 36 algorithm runs took approximately 3.5 hours. To address this issue, increasing the sampling rate to reduce the number of generations included in subsequent stages is a viable solution. Advanced sampling techniques can also be explored to preserve only informative generation subsequences.

\vspace{1.1mm}\noindent\textit{Visual Design}: Our framework utilizes scatterplot-based designs, in the algorithm similarity view, the generation similarity view, and the solution set view. When a large number of elements, i.e., points, are rendered in the same view, visual clutter may occur. Moreover, considering that colors are employed to represent distinct algorithms, there may be difficulties when a considerable number of algorithms are activated in the aforementioned views, as previously recognized in the literature on multi-class scatterplots~\cite{Yuan2021Evaluation,Lu2023multiclass}. Our framework has deployed various techniques such as semantic zooming and density maps to mitigate the issues. Subsequent work could explore additional simplification strategies, such as hierarchical visualization techniques, additional filtering methods, and multi-class sampling approaches for scatterplots.

\vspace{1.4mm}\noindent\textbf{Generalizability.} First, with regard to the number of objectives, our framework can be readily extended to address many-objective optimization problems where the number of objectives exceeds that of the multi-objective setting. This can be achieved through the incorporation of advanced multi-dimensional projection methods, such as t-SNE and UMAP, into the visualization of solution sets. Second, a promising extension for generalizing our framework lies in the realm of stochastic behaviors in multiple runs of an algorithm. Here, additional uncertainty analysis and visualization techniques can be leveraged to enhance the analysis of such behaviors. In addition, our framework can be employed in a wide range of iterative optimization algorithms, where iterations can be treated as generations in the EMO context. Nonetheless, significant modifications to the solution visualization components may be necessary to support the analysis of only one objective.

\vspace{1.4mm}\noindent\textbf{Limitations and Future Work.} Currently, our framework adopts an abstract representation of the objective space. In the future, we intend to enhance our framework for specific application scenarios by incorporating contextual information and semantic meanings into the visualizations. We also aim to uncover the underlying mechanisms of the evolutionary processes by applying white-box explanation techniques to EMO algorithms. Furthermore, there are current plans to enhance the visual exploration components for the solution sets by adopting additional multi-dimensional visualization methods. A linked analysis between the decision and the objective space can significantly facilitate the understanding of solution distributions in a single solution set. Lastly, there remain several parameters that users can adjust, including the down-sampling rate of the algorithm runs and the $k$ value for the kNN graph. Further investigation can be undertaken to understand the impact of these hyperparameters on the analytical workflow.

%% if specified like this the section will be ommitted in review mode

\section*{Supplemental Materials}
\label{sec:supplemental_materials}

%All supplemental materials are available on OSF at \url{https://doi.org/10.17605/OSF.IO/2NBSG}, released under a CC BY 4.0 license.
The supplemental materials include (1) a demo video of our framework, (2) the subtitle file of the demo video, and (3) the list of all EMO algorithms selected in the case studies. An implementation is released at~\url{https://github.com/VIS-SUSTech/visual-analytics-for-emo-algorithm-comparison}.

\acknowledgments{%
  This work was supported in part by the National Natural Science Foundation of China (No. 62202217), Guangdong Basic and Applied Basic Research Foundation (No. 2023A1515012889), Guangdong Talent Program (No. 2021QN02X794), and the Program for Guangdong Introducing Innovative and Entrepreneurial Teams (No. 2017ZT07X386). Y. Ma would like to thank his wife, Y. Qi, for her love and constant support.
}

\bibliographystyle{abbrv-doi-hyperref}

\bibliography{template}

%% ^^^^^   FOR IEEE VIS, EVERYTHING HERE MAY BE INCLUDED IN THE    ^^^^^ %%
%% 2-PAGE ALLOTMENT FOR REFERENCES, FIGURE CREDITS, AND ACKNOWLEDGEMENTS %%

%\appendix % You can use the `hideappendix` class option to skip everything after \appendix

\end{document}